\documentclass[times,twocolumn,final]{elsarticle}

\usepackage{graphicx}
\graphicspath{{./figs/}}
\DeclareGraphicsExtensions{.pdf}

\usepackage{amssymb}
\usepackage[cmex10]{amsmath}

\usepackage{url}
\usepackage{epstopdf}
\usepackage{pdfpages}
\usepackage{colortbl}
\usepackage[ruled,vlined]{algorithm2e}

\journal{}

\begin{document}

\begin{frontmatter}



\title{Iterative Universal Hash Function Generator for Minhashing}


\author[olivetti]{Fabr\'icio~Olivetti~de~Fran\c{c}a}
\ead{olivetti@ieee.org}
\address[olivetti]{Center of Mathematics, Computing and Cognition (CMCC), Universidade Federal do ABC (UFABC) -- Santo Andr\'{e}, SP, Brazil.}

\begin{abstract}
Minhashing is a technique used to estimate the Jaccard Index between two sets by exploiting the probability of collision in a random permutation. In order to speed up the computation, a random permutation can be approximated by using an universal hash function such as the $h_{a,b}$ function proposed by Carter and Wegman. A better estimate of the Jaccard Index can be achieved by using many of these hash functions, created at random. In this paper a new iterative procedure to generate a set of $h_{a,b}$ functions is devised that eliminates the need for a list of random values and avoid the multiplication operation during the calculation. The properties of the generated hash functions remains that of an universal hash function family. This is possible due to the random nature of features occurrence on sparse datasets. Results show that the uniformity of hashing the features is maintaned while obtaining a speed up of up to $1.38$ compared to the traditional approach.
\end{abstract}


\end{frontmatter}


\section{Introduction}
\label{sec:intro}

Many tasks in Machine Learning require the calculation of similarity for every pair of objects in the data set. Some examples of such tasks are data clustering and nearest neighbors searching. The need for such calculation may be demanding for large data sets with many features leading to the so called curse of dimensionality~\cite{friedman1997bias}. This occurs when the objects of a data set are described on a higher dimension, typically with hundreds of features, adding one more dimension to the computational complexity, i.e., a $O(n)$ becomes $O(n.d)$, where $n$ is the number of objects and $d$ the number of features. Also, a higher dimension may also lead to a loss of precision when calculating the similarities between two object since this may only be perceived on a small subset of features and thus masked by all the others, that could be considered random noise.

One way to deal with such problem is to transform the data into a low dimension representation while preserving enough information in way that similar objects remains similar after the transformation. Some methods used to reduce the dimensionality of a data set are Feature Selection~\cite{guyon2003introduction}, Feature Extraction~\cite{guyon2006feature} and Probabilistic Dimension Reduction~\cite{lawrence2008dimensionality}. 

The Feature Selection approach tries to find the subset of the features that are most relevant to the task being performed. This can be done by removing redundant (highly correlated) variables and irrelevant variables that brings no information whatsoever. These are usually measured by the Entropy, Information Gain and Correlation measures. Since this is a combinatorial problem, it is mostly solved with meta-heuristic approaches~\cite{castro2008feature,vieira2012metaheuristics}.

Feature Extraction refers to the creation of a smaller features set based on the (non-)linear combination of the original features. This is usually performed by using matrix decomposition techniques, creating a map between the original features set to the transformed one. More recently there has been a focus on Neural Network Feature Extraction through Deep Learning techniques~\cite{bengio2012unsupervised}.

Probabilistic Dimension Reduction is performed exploiting the probability of two objects being very similar. This is done by means of specially designed hash functions that has a high probability of collision when hashing objects with similarity above a certain threshold. One of these techniques is called Locality-Sensitive Hashing~\cite{har2012approximate,andoni2013beyond} and was successfuly applied to different applications including text and image retrieval. The main advantage of this technique is that the computational complexity approximates $O(n)$.

One of such technique is called Minhashing~\cite{666900}, and it was specifically crafted to approximate the Jaccard Index between two sets. This is based on the fact that, given a random permutation of the features set, the probability that the first feature of two objects are the same is equal to the Jaccard Index between those two. In order to become computationally feasible, the random permutation is approximated by using an universal hash function.

Since a good estimation through probabilistic procedures require a sample of considerable size, there is a need to automatically create universal hash functions at random. One of such approaches, as devised by Carter and Wegman~\cite{Carter1979143}, requires that two random values are drawn from an uniform distribution for each function. Given that $N$ hash functions are created, there will be a total of $2N$ random values. This leads to a requirement of additional $O(2N)$ space complexity and $N$ multiplication operations.

The need for an optimization is justified since this procedure is often used on very large datasets~\cite{szmit2013locality} and with algorithms that requires a very large number of those hash functions~\cite{franca2012scalable} where a small speed up might imply on an economy of hours of processing.

In this paper it will be described an iterative procedure for generating any number of hash functions without the need of generating random numbers. This is done by exploiting the randomness of the features occurrence on sparse data sets. The speed up of this method will be experimently measured as well as the uniformity properties. In Section~\ref{sec:minhashing} it will be explained how minahshing works and the general algorithm. Section~\ref{sec:iterative} will explain the proposed modifications, describe the pseudo-algorithm and show some initiall results regarding the uniformity of this approach. The performance and properties of such modification will be analysed and explored during the experimental setup of Section~\ref{sec:results}. Finally, in Section~\ref{sec:conclusion} it will be given some concluding remarks.

\section{Min-wise Independent Permutations Locality Sensitive Hashing}  
\label{sec:minhashing}

When the features of the studied data set can be described as sets, as it is the case on many semi-structure data (i.e., text documents), the similarity between two objects can be calculated through the Jaccard Index or Jaccard similarity:

\begin{equation}
  J = \frac{\left|O1 \cap O2\right|}{\left|O1 \cup O2\right|},
\end{equation}

\noindent where $O1$ and $O2$ are the two objects compared and it returns a value between $0$ and $1$, with the latter meaning the two objects are equal.

Let us illustrate this concept by using an example of comparing two text documents:

\begin{center}
d1 = "The cat ate the mouse"\\
d2 = "The mouse ate the cheese"
\end{center}

One data strcuture used to represent these documents is an ordered set of terms contained in a document, or the bag-of-words representations:

\begin{center}
d1 = \{ate, cat, mouse, the\}\\
d2 = \{ate, cheese, mouse, the\}
\end{center}

The Jaccard similarity between these two sets will be:

\begin{equation}
J(d1,d2) = \frac{\left|d1 \cap d2 \right|}{\left|d1 \cup d2\right|} = \frac{3}{5}.
\end{equation}

Since the intersection and union operations can be costly to compute, it can become very time consuming to perform pairwise similarity computation on a large data set with objects represented as a large set of features. As such, it is desirable to find a way to estimate the Jaccard index and reduce the objects representation to a smaller dimension.

The technique known as Min-wise independent permutations locality sensitive hashing (MinHash)~\cite{har2012approximate}, allows to quickly estimate the similarity of two sets approximating the Jaccard Index. It is based on the probability that, given a random permutation of two sets, the first element of both sets will be the same with a probability equal to their Jaccard Index.

Following the same example, suppose the ordered set of terms is shuffled following a permutation $\pi$ and the documents are represented by:

\begin{center}
d1 = \{the, mouse, cat, ate\}\\
d2 = \{the, mouse, cheese, ate\}
\end{center}

The probability that the first element of these sets are the same is equal to the ratio between the number of common elements and the number of total distinct elements, or:

\begin{equation}
P( d1_0^\pi = d2_0^\pi ) = \frac{\left|d1 \cap d2 \right|}{\left|d1 \cup d2\right|} = \frac{3}{5} = J(d1,d2).
\end{equation}

This probability can be estimated by reshuffling the ordered set with $N$ different permutations and counting the number $N_{equal}$ of times that the first element are equal to both sets:

\begin{equation}
P( d1_0^\pi = d2_0^\pi ) \approx \frac{N_{equal}}{N} \approx J(d1,d2).
\end{equation}
 
So, it is possible to estimate the Jaccard Index between two objects, described by a total of $M$ features, by calculating $N$ Minhashes, with $N$ different permutations, for each document. Notice that the permutation must follow an uniform distribution.

Given that $N < M$, the cost of comparing two documents is reduced by a factor of $\frac{N}{M}$. Since the cost of generating a permutations is $O(M)$, the total computational cost will be $O(N.(M + D.\bar{M}))$ where $D$ is the number of documents and $\bar{M}$ is the average number of features per document. One way to avoid this additional cost is by using an universal hash function such as one of those proposed in~\cite{Carter1979143}:

\begin{equation}
h(x) = a \cdot x + b \mod P,
\label{eq:h}
\end{equation}

\noindent where $a$ and $b$ are randomly chosen with uniform distribution, and $P$ is a large prime number. The variable $x$ is the value to be hashed, i.e., a number associated with a given feature. The prime number should be at least as large as the number of features. This hash function will map each feature to an index in the range $[0,P[$. The random values will ensure that those indeces generate a random permutation of the feature set.

Notice that with this function the application of MinHash is straightforward. For each object $j$, simply find, for each hash function $i$, the feature $x$ that has the minimum value for $h_i(x)$:

\begin{equation}
mh_i(o_j) = \underset{x}{\arg\min} \left\{h_i(x) \mid \forall  x \in o_j\right\}
\end{equation}

Since the probability that the minimum indices of two objects are the same, given a hash function, equals the Jaccard Index as well The complexity with this approach becomes $O(N.D.\bar{M})$ with usually $\bar{M} << M$ on sparse data sets. 





\section{Iterative Procedure}
\label{sec:iterative}

By inspecting Eq.~\ref{eq:h}, the purpose of the random values $a$ and $b$ is to ensure uniformity of permutation given a sequence of features indeces. In other words, if we have $M$ features, in $M$ different permutations, each feature is expected to have the minimum hash value once.

In many real world applications the set of features describing each object is sparse (small percentage of the full feature set) and presents a randomness following a probability distribution. The randomness of features occurrence can be exploited to avoid the random values calculation. Let us first define the $i$-th hash function, of a sequence of $N$ functions, as:

\begin{equation}
h_{i} = (a+i) \cdot x  + i \cdot b \mod P.
\end{equation}

Notice that this is still the same universal hash function as before, holding the same properties. If we subtract $h_{i-1}$ from $h_{i}$ in modulus $P$, we obtain:

\begin{align*}
\Delta h &= h_{i} - h_{i-1} \\
		&= \left[ (a+i) \cdot x  + i \cdot b \mod P \right.\\
		&- \left. (a+i-1) \cdot x + (i-1) \cdot b \mod P \right] \\
		&= (x + b) \mod P,
\end{align*}

\noindent when $P > a,b$.

So, given a value $x$ to be hashed, and the value of the first hash function $h_{0}(x)$, we can obtain any number of hash functions by sequentially summing up $\Delta h$. This procedure is described in Alg.~\ref{alg:iterativehash}.

\begin{algorithm}
\KwData{the value $x$ to be hashed, initial values for $a$ and $b$, a large prime number $P$ and the number $N$ of hash functions.}
\KwResult{vector $H$ with the $n$ hash values of $x$}
\BlankLine
$H[0] = a*x$\;
$\Delta h = b+x \mod P$\;
\For{$i\leftarrow 1$ \KwTo $N$}{
	$H[i] = H[i-1] + \Delta h \mod P$\;
}
\label{alg:iterativehash}
\caption{Iterative universal hash function generator.}
\end{algorithm}

The main difference between the random approach and the iterative one is that the latter avoid the multiplication operation. This reduce the computational complexity in about $O(N \cdot b \cdot log b \cdot log log b)$ (but this varies depending on the multiplication algorithm used). Also, as stated before, the space complexity is reduced from $O(2N)$ to $O(2)$ regarding the list of random numbers.

Notice that this iterative procedure can also be simplified to a function $H(i) = a \cdot x + i \cdot \Delta h \mod P$, so we can still paralellize the calculation of each $H[i]$ if so is required. Additionally, when the algorithm is run on a distributed framework (i.e., MapReduce) the overhead of sending a vector of random numbers of size $2N$ throughout all of the machines is eliminated. If not for a significant speedup, then for a concise and clearer code.

Since the rationale for the proof of universality of this hash function is the same as the random approach, in the next section it will be provided some empirical experiments in order to test the validity of these claims. It will be compared, between the random and the iterative approach, the uniformity of buckets distribution, the uniformity of choosing each feature independetly from the Minhashing procedure, the Jaccard estimation error and the computational time to generate a set of either hash functions.

\section{Experimental Results} 
\label{sec:results}

This section will be devided into two subsections: the first one will test the uniformity of distribution of hashed values to $m$ buckets and that the probabiliy of a value $x$ be chosen as a Minahsh is uniform; the second will show that the estimation error of the Jaccard Index is similar between both approaches with a slightly advantage for the iterative proceudre and, finally, the obtained speed up.

\subsection{Uniformity of Distribution}

The uniformity of an universal hash function states that, given a hash function with $m$ buckets, the probability of a value $x$ being assigned to a bucket with a random hash function $h$ is proportional to ${}^{1}/_{m}$. To verify if the family of Iterative Hash functions has this property, it was performed $100$ independent experiments with a different value $x$ randomly assigned in the range $[0,P[$, where $P = 7,757$ was the prime number chosen, to be hashed in one of $100$ different buckets. To evaluate the uniformity it was performed a $\chi$-squared test for each experiment with $\alpha = 0.05$ with $1,000$ hash functions. The choice for this number of hash functions is due to a recomendation that the expected count for each bucket should be $5$ to $10$ in order for the $\chi$-squared test to return an acurate response. 

Similarly, for each experiment it was generated $100$ random key values and the Minhash for each of the $1,000$ hash functions was calculated. The uniformity of the probability that a given key will be chosen as the Minhash was also tested with the $\chi$-squared test with $\alpha = 0.05$. Table~\ref{tab:resultUniform} reports the percentage of the $100$ experiments where the $\chi$-squared test pointed to a likely uniformity.

\begin{table}
\centering
\caption{Percentage of likely results from a $\chi$-squared Test for the uniformity of distribution of the Random and Iterative Hash function with $1,000$ hashes and $100$ buckets and features.}
\begin{tabular}{c|c|c}
\hline
& \textbf{Random} & \textbf{Iterative} \\
\hline\hline
\textbf{Uniformity Test} & $92\%$ & $92\%$ \\
\textbf{Minhashing Test} & $95\%$ & $93\%$ \\
\hline
\end{tabular}
\label{tab:resultUniform}
\end{table}

From these results we can see that both family of hash functions are likely to generate an uniform distribution of buckets and a uniform distribution of choice for the Minhash. In order to simulate a more realistic scenario, where we can have thousands of possible features (i.e., text mining), the same tests were performed but with $5,000$ buckets and features and $50,000$ hash functions. The results are reported on Table~\ref{tab:resultUniformBig}.

\begin{table}
\centering
\caption{Percentage of likely results from a $\chi$-squared Test for the uniformity of distribution of the Random and Iterative Hash function with $50,000$ hashes and $5,000$ buckets and features.}
\begin{tabular}{c|c|c}
\hline
& \textbf{Random} & \textbf{Iterative} \\
\hline\hline
\textbf{Uniformity Test} & $100\%$ & $100\%$ \\
\textbf{Minhashing Test} & $93\%$ & $100\%$ \\
\hline
\end{tabular}
\label{tab:resultUniformBig}
\end{table}

As we can see, these results just confirms the previous experiment that both set of hash functions likely have the uniformity property and belongs to the universal family of hash functions.

\subsection{Hash Function Performance}

Next the processing time for both approaches were compared in order to quantify the speedup obtained with the iterative hash function. For this end it was performed $100$ independent experiments to generate and apply $100,000$ hash functions for a subset of $500$ documents from the $20$ Newsgroups dataset~\cite{Lang95} and $10,000$ hash functions for a subset of $3,891$ documents from the Classic dataset~\cite{classic3}. This code was written in C++ with Boost Library and compiled with g++ 4.7.3 using just the optmization flag \textit{-O2} on an Intel i5 2.5GHz using a single core. The Operational System used was the Mint 15 Linux Distribution and the source code for all of the reported experiments can be found at \url{https://github.com/folivetti/HBLCoClust/}.

\begin{table}
\centering
\caption{Average and standard deviation of the time, in seconds, obtained for each experiment.}
\begin{tabular}{c|c|c}
\hline
&\textbf{Random} & \textbf{Iterative} \\
\hline\hline
$20$-Newsgroup & $26.27 \pm 1.59$ & $18.94 \pm 0.56$ \\
Classic & $12.61 \pm 0.24$ & $10.10 \pm 0.20$ \\
\hline
\end{tabular}
\label{tab:resultTime}
\end{table}

In Table~\ref{tab:resultTime} the average and standard-deviation of the time taken by every experiments are reported, it is possible to see that there is a speedup of $1.38$, on average, for the $20$-Newsgroup dataset and $1.25$ for the Classic dataset, also a t-paired test was performed with $\alpha = 0.05$ which confirmed that the difference is statistically significant for both experiments. As the results point out, this is a significant improvement on the computational time that can chiefly benefit the computation of datasets with millions of objects. If, for example, an experiment using the traditional random approach takes $10$ hours on a very large dataset, with the iterative procedure we can expect about $7$ hours of processing.

Finally, in order to see if the Jaccard coefficient estimation of both hash approaches are equivalent, the Minhashing estimation was performed on the same dataset with each object being represented by ${5,10,15}$ hashes and the Mean Absolute Error was calculate between each estimation and the real Jaccard Index.

\begin{table}
\centering
\caption{Mean Absolute Error of the estimations obtained by the Random and the Iterative approaches.}
\begin{tabular}{c|c|c}
\hline
\textbf{\# of hashes} & \textbf{Random} & \textbf{Iterative}\\
\hline\hline
$5$ & $0.0455 \pm 0.0368$ & $0.0456 \pm 0.0368$ \\
$10$ & $0.0430 \pm 0.0273$ & $0.0421 \pm 0.0255$ \\
$15$ & $0.03859 \pm 0.0231$ & $0.0387 \pm 0.0230$ \\
\hline
\end{tabular}
\label{tab:resultEstimation}
\end{table}

The results from Table~\ref{tab:resultEstimation} shows a small difference of the mean absolute error obtained by each approach. It should be noticed that a t-test performed on each experiment revealed that the p-value for the experiments with $10$ and $15$ hashes were below $0.05$ which means that, althought just slightly different, the means are unlikely the same. The p-value obtained on the $5$ hashes experiments was much higher than $0.05$, though.

\section{Conclusion} 
\label{sec:conclusion}

This paper proposed an optimization to the random universal hash function generator commonly used to estimate the Jaccard Coefficient between two sets. The proposal eliminates the need of generating two large lists of random values and the repetition of a multiplication operation throughout the calculation. This is done by changing the function creation to an iterative procedure where a given hash function $h_i(x)$ depends on the previously generated function plus an increment.

As a first concern, it was tested whether this iterative procedure keeps the universality properties of the random hash functions. Some experiments indicates that this hash function has uniformity of buckets allocation and Minhashing distribution, as such, it can be used as an universal hash function.

Next, the speed up obtained with this procedure was measured and the average gain was between $1.25$ to $1.38$ from the time measured with the random approach. Additionally, it was shown, with another experiment, that the estimation error regarding the Jaccard Index, remained pratically the same, with a tiny advantage to the iterative approach.

These results show that the iterative procedure can have a significant impact when mining very large datasets and when using hundreds of hash functions. And, since it is just a simple modification, it can be easily adapted onto many existing source codes that already uses such hash functions.





\bibliographystyle{elsarticle-num}
\bibliography{IterativeMinHashing}







\end{document}